\pdfoutput=1

\documentclass[11pt]{article}

\usepackage[preprint]{acl}

\usepackage{times}
\usepackage{latexsym}

\usepackage[T1]{fontenc}

\usepackage[utf8]{inputenc}

\usepackage{microtype}

\usepackage{inconsolata}

\usepackage{graphicx}
\usepackage{multirow}
\usepackage{makecell}

\usepackage{subcaption}
\usepackage{booktabs}

\title{Analyzing Dialectical Biases in LLMs for Knowledge and Reasoning Benchmarks
}

\author{
  \textbf{Eileen Pan\textsuperscript{1}},
  \textbf{Anna Seo Gyeong Choi\textsuperscript{1}},
  \textbf{Maartje ter Hoeve\textsuperscript{2}},
  \textbf{Skyler Seto\textsuperscript{2}},
  \textbf{Allison Koenecke\textsuperscript{1,}\textsuperscript{3}}
\\
\\
  \textsuperscript{1}Department of Information Science, Cornell University, Ithaca, NY \\
  \textsuperscript{2}Apple, Cupertino, CA \\
  \textsuperscript{3}Cornell Tech, New York, NY
\\
  \small{
    \textbf{Correspondence:} \href{mailto:koenecke@cornell.edu}{koenecke@cornell.edu},\href{mailto:sseto@apple.com}{sseto@apple.com}
  }
}

\begin{document}
\maketitle
\def\thefootnote{*}
\begin{abstract}
Large language models (LLMs) are ubiquitous in modern day natural language processing. However, previous work has shown degraded LLM performance for under-represented English dialects. 
We analyze the effects of typifying ``standard'' American English language questions as non-``standard'' dialectal variants on multiple choice question answering tasks and find up to a 20\% reduction in accuracy. 
Additionally, we investigate the grammatical basis of under-performance in non-``standard'' English questions. 
We find that individual grammatical rules have varied effects on performance, but some are more consequential than others: three specific grammar rules (existential ``it'', zero copula, and y'all) can explain the majority of performance degradation observed in multiple dialects. We call for future work to investigate bias mitigation methods focused on individual, high-impact grammatical structures.  

\end{abstract}

\section{Introduction}

Large language models are essential to natural language processing applications, achieving strong performance across numerous tasks. 
However, language model learning is highly sensitive to the style of the data used in training~\cite{maini2024rephrasing}. This can lead to fairness issues for speakers of non-``standard'' varieties of American English (such as African American English, Chicano English, and non-native English speakers), who may write text in their corresponding spoken dialects~\cite{whiteman2013dialect,smitherman1986talkin,harvey2025framework,blodgett_demographic_2016,johnson2012linguistic, hofmann2024ai}; in turn, these written variants are likely underrepresented in training corpora relative to ``Standard American English'' (SAE). These are particularly important English-speaking populations to study as they have already been found to suffer from LLM underperformance in benchmark tasks~\cite{ryan2024unintended,hofmann2024ai}, and are often correspondingly minoritized in broader societal contexts. An example of differential validity on such tasks is if LLMs disproportionately respond with inaccurate responses to prompts written in African American English, but respond correctly to prompts written in SAE. This is a real concern, as LLMs are increasingly involved in high-stakes scenarios from education to hiring.

While LLM underperformance of individual varieties of English has been studied at a high level in real-world conversational contexts \cite{ziems2022multi, srirag-etal-2025-evaluating}, it remains understudied the extent to which (a) underperformance for different English dialects is an issue in more basic NLP tasks (i.e., multiple choice rather than open response questions), and (b) grammatical rules defining English dialects might be drivers of LLM response differences. These questions have historically been difficult to answer because one would need to find corpora of comparable text typified in different dialects; however, prior work automating dialectal translation~\cite{ziems2022value, ziems2022multi} allows us to generate text data in different dialects based on grammatical rules. We apply this translational tool to fill a gap in the literature by answering two research questions:

\textbf{RQ1}: Do LLMs underperform when answering multiple choice questions that are typed in a written dialect (African American English, Appalachian English, Chicano English, Indian English, Singaporean English, and Southern English) versus answering questions typed in SAE?

\textbf{RQ2}: Can LLM underperformance in certain dialects be decomposed into underperformance stemming from multiple specific grammatical rules?

Answering these questions is important: not only can we quantify biases in fundamental LLM tasks, but we can further break down these quantities by grammatical rules, which can inform model developers of whether these rules should be a focus of improvement for multidialectal LLMs, thereby helping to drive mitigation of identified biases. 

Prior work addressing these questions has either focused on linguistic analyses of model underperformance for individual grammatical rules, such as the \emph{habitual be}~\cite{martin2020understanding} or \emph{zero copula}~\cite{koenecke2020racial} common to African American English~\cite{rickford2007spoken}, or has focused on studying biases in overall dialects without considering individual grammatical rules~\cite{lin2024one,hofmann2024ai,srirag-etal-2025-evaluating}. In contrast, we study a wide range of grammatical rules used across multiple dialects, which can in turn be used to inform model improvements across multiple dialects: given the high number of shared or similar grammatical features across dialects, we may expect that technical improvements on specific grammatical rules can yield performance improvements across dialects through transfer learning.

\begin{table*}[ht]
\centering
\tiny
\setlength{\tabcolsep}{3pt} %
\begin{tabular}{l|cccc|cccc|cccc|}
\hline
\multirow{2}{*}{English Variety} & \multicolumn{3}{c|}{BoolQ Accuracy (\%)}  & Count & \multicolumn{3}{c|}{SciQ Accuracy (\%)} & Count & \multicolumn{3}{c|}{MMLU Accuracy (\%)}  & Count  \\
\cline{2-13}

 & Gemma 2B & Mistral 7B & GPT4o-mini & 
 & Gemma 2B & Mistral 7B & GPT4o-mini &
 & Gemma 2B & Mistral 7B & GPT4o-mini & \\
\hline
Standard American English 
& 71.3 & 85.4 & 87.7 & 9348
& 94.3 & 96.6 & 97.7 & 11647
& 34.1 & 61.3 & 71.5 & 11672\\
Chicano English     
& 72.0 (+0.7) & 84.5 (-0.9) & 87.2 (-0.5) & 3656 
& 93.8 (-0.5) & 96.6 (0.0) & 97.6 (-0.1) & 4325 
& 32.9 (-1.2) & 56.9 (-4.4) & 65.4 (-6.1) & 6772 \\
Appalachian English     
& 70.9 (-0.4) & 83.2 (-2.2) & 85.7 (-2.0) & 5725 
& 93.4 (-0.9) & 96.5 (-0.1) & 97.2 (-0.5) & 9842 
& 34.9 (+0.8) & 59.6 (-1.7) & 68.7 (-2.8) & 9776 \\
Southern English    
& 69.2 (-2.1) & 83.3 (-2.1) & 85.8 (-1.9) & 8280 
& 93.5 (-0.8) & 96.3 (-0.3) & 97.2 (-0.5) & 11351 
& 34.0 (-0.1) & 59.5 (-1.8) & 69.8 (-1.7) & 11193 \\
African American English    
& 67.7 (-3.6) & 82.6 (-2.8) & 85.8 (-1.9) & 9077 
& 93.3 (-1.0) & 96.1 (-0.5) & 97.0 (-0.7) & 11445 
& 34.2 (+0.1) & 59.5 (-1.8) & 69.2 (-2.3) & 11182 \\
Indian English     
& 68.1 (-3.2) & 81.2 (-4.2) & 85.4 (-2.3) & 9321 
& 92.7 (-1.6) & 95.8 (-0.8) & 96.5 (-1.2) & 11631 
& 34.7 (+0.6) & 59.4 (-1.9) & 68.8 (-2.7) & 11554 \\
Singaporean English  
& 66.5 (-4.8) & 79.8 (-5.6) & 84.6 (-3.1) & 9323 
& 91.7 (-2.6) & 94.7 (-1.9) & 96.1 (-1.6) & 11642 
& 34.1 (0.0) & 58.5 (-2.8) & 68.7 (-2.8) & 11612 \\

\hline
\end{tabular}
\caption{Performance comparison across English varieties with unperturbed questions excluded. In nearly all cases, performance is worse for non-SAE varieties of English.}
\label{tab:dialect_accuracy_only_applied}
\end{table*}

\begin{table*}[ht!]
\centering
\scriptsize
\setlength{\tabcolsep}{3pt} %
\begin{tabular}{l|ccc|ccc|ccc|}
\hline
\multirow{2}{*}{English Variety} & \multicolumn{3}{c|}{BoolQ Accuracy (\%)}  & \multicolumn{3}{c|}{SciQ Accuracy (\%)} & \multicolumn{3}{c|}{MMLU Accuracy (\%)} \\
\cline{2-10}
 & Gemma 2B & Mistral 7B & GPT4o-mini 
 & Gemma 2B & Mistral 7B & GPT4o-mini 
 & Gemma 2B & Mistral 7B & GPT4o-mini \\
\hline
Standard American English 
& 100 & 100 & 100 
& 100 & 100 & 100 
& 100 & 100 & 100 \\
Chicano English     
& 93.9 (-6.1) & 95.6 (-4.4) & 96.7 (-3.3) 
& 99.2 (-0.8) & 99.6 (-0.4) & 99.5 (-0.5) 
& 89.3 (-10.7) & 92.9 (-7.1) & 95.2 (-4.8) \\
Appalachian English     
& 92.0 (-8.0) & 93.6 (-6.4) & 94.8 (-5.2) 
& 98.1 (-1.9) & 99.0 (-1.0) & 99.2 (-0.8) 
& 86.8 (-13.2) & 93.0 (-7.0) & 93.8 (-6.2) \\
Southern English    
& 90.1 (-9.9) & 93.1 (-6.9) & 94.8 (-5.2) 
& 98.4 (-1.6) & 99.1 (-0.9) & 98.9 (-1.1) 
& 83.1 (-16.9) & 92.6 (-7.4) & 92.4 (-7.6) \\
African American English    
& 85.9 (-14.1) & 91.9 (-8.1) & 95.0 (-5.0) 
& 98.2 (-1.8) & 99.1 (-0.9) & 98.8 (-1.2) 
& 84.4 (-15.6) & 92.3 (-7.7) & 92.3 (-7.7) \\
Indian English     
& 86.9 (-13.1) & 90.2 (-9.8) & 93.6 (-6.4) 
& 97.5 (-2.5) & 98.4 (-1.6) & 98.5 (-1.5) 
& 81.3 (-18.7) & 91.2 (-8.8) & 90.8 (-9.2) \\
Singaporean English  
& 83.3 (-16.7) & 88.2 (-11.8) & 92.3 (-7.7) 
& 96.4 (-3.6) & 98.0 (-2.0) & 97.4 (-2.6) 
& 78.4 (-21.6) & 89.9 (-10.1) & 88.8 (-11.2) \\
\hline
\end{tabular}
\caption{Performance comparison across English varieties with unperturbed questions excluded, conditioned on correct answers in Standard American English. In all cases, performance is worse for non-SAE varieties of English.}
\label{tab:dialect_accuracy_only_applied_sae_correct}
\end{table*}

\begin{figure*}[ht]
  \includegraphics[width=\linewidth]{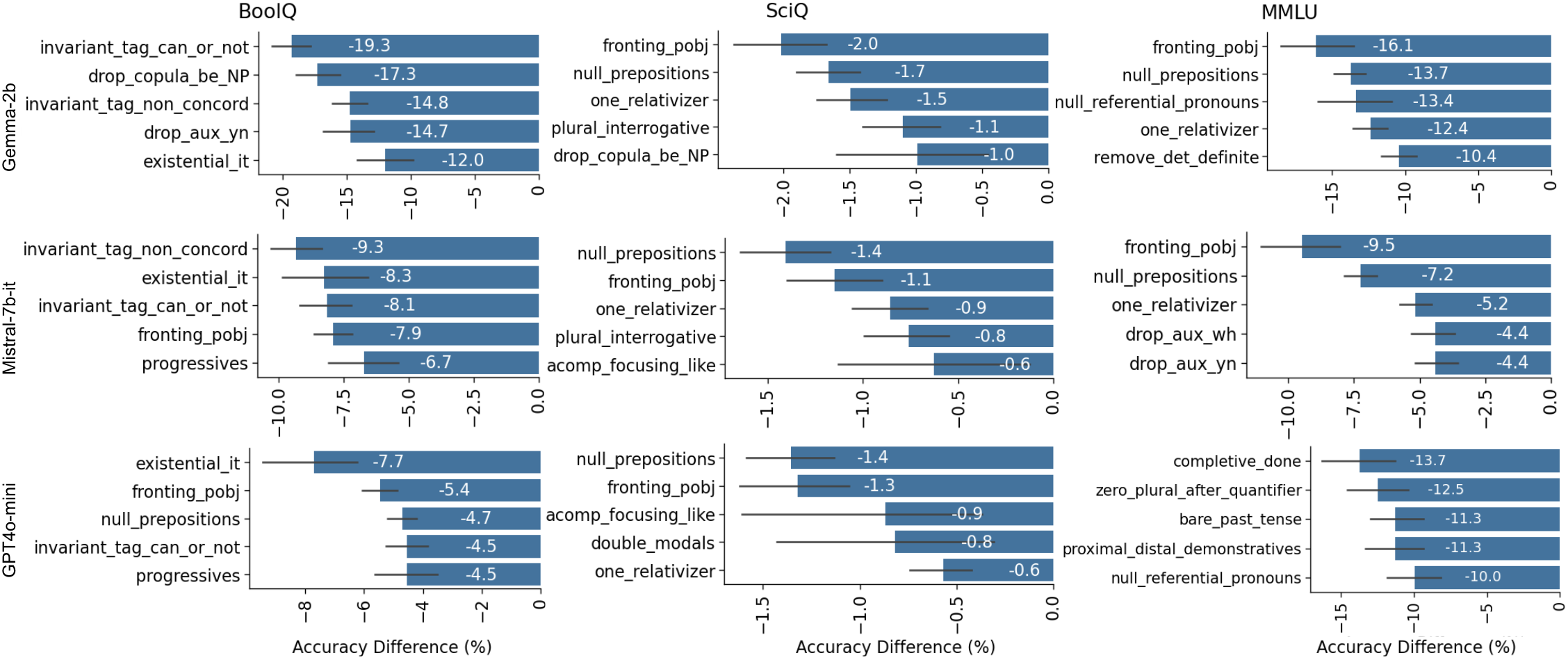}
  \caption {Grammatical rules are applied to full QA datasets one at a time; the top five rules by accuracy reduction are shown for each facet (on the subset of questions for which the grammatical rule can be applied, and for which the LLM answered correctly when asked in SAE). Accuracy difference refers to the comparison between the original QA dataset and the QA dataset having applied only a single grammatical rule. Grammatical rule definitions are provided in Table~\ref{tab:grammar_table}.}
  \label{fig:grammar_bars}
\end{figure*}

\section{Methods}

We begin with three multiple choice benchmark Question Answering (QA) datasets commonly used for benchmarking: BoolQ~\cite{clark2019boolq} containing 9,427 real Google user queries, SciQ~\cite{welbl2017crowdsourcing} containing 11,679 science exam questions, and MMLU~\cite{hendrycks2020measuring} containing 14,042 questions spanning 57 subtopics from accounting to religion. All datasets contain both questions and multiple choice answers in ``Standard American English" (SAE). We opt to focus on multiple choice questions as we expect these tasks to be relatively easy for LLMs in SAE; uncovering dialectal underperformance in these tasks could indicate the extent to which dialectal biases persist across not just difficult---but also prima facie easier---tasks, and can potentially serve as a lower bound on quality-of-service harms across LLM question answering tasks.

Next, we generate grammatically-perturbed variants of each question in these datasets---leaving answers and support material unchanged---by using the Multi-VALUE package~\cite{ziems2022multi}. Multi-VALUE is based on the Electronic World Atlas of
Varieties of English (eWAVE), a linguistic database of morphosyntactic variation in spontaneous spoken English~\cite{ewave}, and allows users to input SAE text and generate dialectal variants based on sets of grammatical rules from eWAVE. The creators of Multi-VALUE additionally recruited speakers of several dialects (including African American English, Chicano English, Indian English, and Appalachian English) to validate Multi-VALUE dialectal translations~\cite{ziems2022multi}.

This package allows us to perturb SAE text (a) by individual grammar rules, (b) by multiple grammar rules of our choosing, and (c) using the default set of grammar rules ascribed to specific dialects (e.g. Appalachian, Singaporean, etc.). For example, for the sentence ``She is always studying,'' we could apply the specific \emph{zero copula} grammatical rule to obtain the string ``She always studying,'' or apply all rules for the African American English dialect (in this case, both \emph{zero copula} and \emph{habitual be}) to obtain the string ``She always be studying.'' Multi-VALUE applies rules probabilistically based on documented dialect pervasiveness, with uncommon dialectal grammar rules up-weighted for stress-testing \cite{ziems2022multi}.

For the six dialects of interest in our study, we use the Multi-VALUE default grammatical rules and transformation frequencies when generating dialectal perturbations. For individual and groups of grammar rules, we set the transformation frequencies to 100\% such that grammar rule transformations are always applied when the corresponding grammatical structure appears. We perturb only question texts and not reference or answer texts. We include examples of individual rule transformations in Table~\ref{tab:grammar_rule_examples}. We exclude questions that Multi-VALUE cannot process from consideration in SAE results. 

We then compare the performance of three common LLMs on both original (SAE) and grammatically-perturbed variants of the three QA datasets. We choose to use Gemma-2B, Mistral 7B, and GPT4o-mini due to their popularity in real-world applications, and to encapsulate a range of model sizes in our evaluations.\footnote{We estimate the project took around 200 GPU hours on 12-20GB VRAM GPUs.} We used the default prompts in LM Eval Harness \cite{eval-harness} available for these datasets in a zero-shot setting. For all three LLMs, we find performance on the unperturbed QA datasets to be comparable to their technical reports \cite{gemmateam2024gemmaopenmodelsbased, jiang2023mistral7b, openaiGPT4oMiniAdvancing2024}. 

We then compare performance for each of the three datasets, for each of the dialectal and grammatical variants of each dataset, and for each of the three LLMs. To calculate accuracy, we first subset to the set of questions that differ from the original (SAE) dataset by at least one grammatical rule (e.g., the ``y'all'' grammatical rule can only be perturbed for a dialect if the word ``you'' appears in the original question), reflected in Table~\ref{tab:dialect_accuracy_only_applied}. Then, we subset to questions that were answered correctly for that LLM when asked in the original (SAE) dialect, reflected in Table~\ref{tab:dialect_accuracy_only_applied_sae_correct}.  We primarily focus on questions correctly answered in SAE as they highlight a clear quality-of-service gap where LLMs are capable of answering a question in SAE but not in a different dialect. Meanwhile, questions that an LLM cannot answer in SAE and still cannot answer in dialect may be less meaningful when considering dialectal disparity. For robustness, we additionally report average accuracy metrics that include all unperturbed questions in Appendix~\ref{app:other_metric_variants}.

Finally, we quantify bias as the percentage point differential in accuracy (for each LLM, and each dataset) between each variant and the original (SAE) questions.

\section{Results}

\subsection{Dialectal Biases}

We find that---on average, across LLMs and QA datasets---prompting LLMs with questions in non-``Standard'' English dialects results in lower accuracy on multiple choice answers per Table~\ref{tab:dialect_accuracy_only_applied}. 
Conditioned on the ``Standard'' English version being correct (Table~\ref{tab:dialect_accuracy_only_applied_sae_correct}), we find even steeper accuracy drops; this is especially true for Singaporean English and African American English, with accuracy drops relative to SAE ranging from 5-16 percentage points for BoolQ and 7-20 percentage points for MMLU. 
We also find that dialectal degradation varies by model and dataset: on BoolQ, the least degradation is observed from GPT4o-mini, whereas on SciQ, the least degradation is observed from Mistral-7B (both on the full question set and when conditioned on SAE accuracy). We generally observe the highest degradations on across tasks from Gemma-2B. Of the tasks, MMLU was the most difficult for LLMs, particularly Gemma-2B, and as a result has more variable degradation.  

This finding is consistent with expectations based on perplexities. As shown in Table~\ref{tab:perplexity_dialects}, we compare the perplexity of each question in its original SAE form, and on the dialectal variant using the FineWeb model \cite{penedo2024the}.\footnote{We consider this model for perplexity analysis as its training dataset set is public and does not appear to contain such grammatically perturbed text. } This shows substantial increases in perplexity when SAE is transformed into dialectal variants, with Singaporean English demonstrating the most dramatic increases, aligning with Singaporean English showing the most substantial performance degradation in our evaluations.

\begin{figure*}[ht]
  \includegraphics[width=\linewidth]{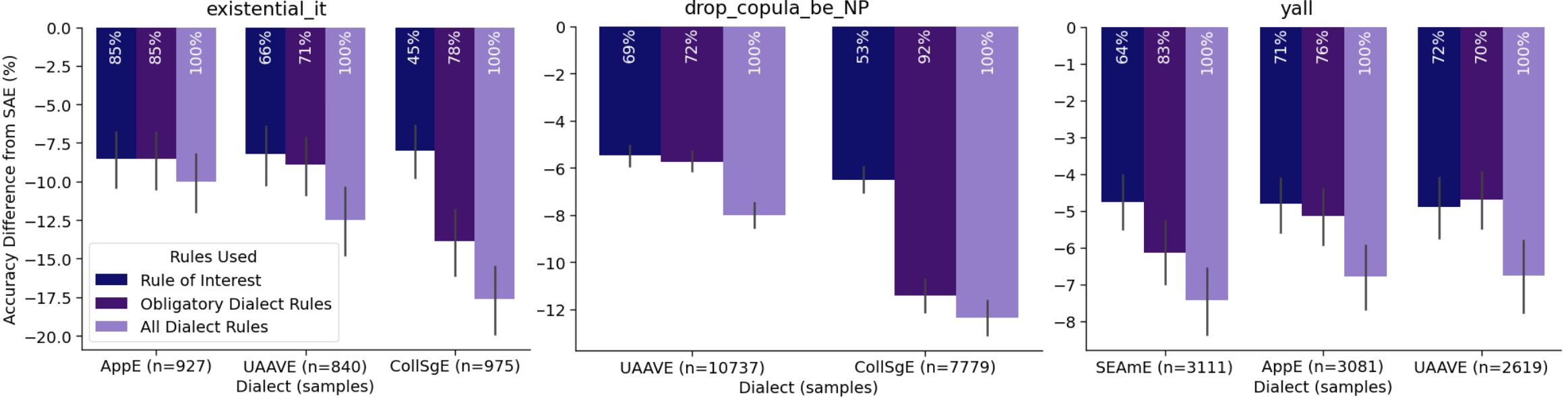}
  \caption {Breakdown of the extent to which accuracy decreases can be attributed to subsets of dialectal grammatical rules. We consider the same data subset--corresponding to the n samples LLMs answered correctly in SAE where the grammar rule is applicable--in each group of bars. We denote the percentage of overall dialectal performance degradation (All Dialect Rules) recovered by just one rule obligatory for that dialect (Rule of Interest) and all rules obligatory (Obligatory Dialect Rules) within the respective bars. Abbreviations are: African American English (UAAVE), Singaporean English (CollSgE), Appalachian English (AppE), and Southern English (SEAmE).}
  \label{fig:grammar_breakdown}
\end{figure*}

\subsection{Grammatical Rule Biases}

We then perform the same analysis at the level of grammatical rule rather than dialectal biases, with a focus on pervasive and representative dialect rules. 
While some grammatical rules could yield a strong decrease in performance, they may be rarely observed for a dialect in practice. As such, we focus on obligatory grammatical rules---rules that are always applied when possible---for the dialects we study that are implemented in Multi-VALUE~\cite{ziems2022multi}. We confirm that the subset of obligatory dialect rules are correlated with overall performance of a dialect, recovering similar trends in degradation as when all dialectal grammar rules are applied. 
For example, we find that the infrequency of obligatory rules in Chicano English is consistent with Chicano English having the least performance degradation among the dialects (see Appendix Figure~\ref{fig:oblig_vs_all}).

We find that obligatory grammatical rules heterogeneously affect QA performance, differing substantially by dataset and model. 
We analyze the accuracy drop from these obligatory grammar rules and highlight the worst performing grammar rules in Figure~\ref{fig:grammar_bars}. Some, such as \emph{remove definite determiner} (dropping ``the''), rarely result in reduced QA task accuracy. Others, such as \emph{fronting} (moving the prepositional phrase to the beginning of the sentence), reduce QA task accuracy in all models. 
Furthermore, we find that some grammar rules significantly reduce accuracy across all LLMs and datasets (p<0.05, McNemar's test);
see Table~\ref{tab:grammar_table} for this list of grammar rules and their occurrences within our corpora. 
We observe that 9 out of 20 of the observed statistically-significant grammar rules overlap across dialects, indicating that future work on mitigating grammar rule-based biases (e.g., training on grammatically-altered variants of QA pairs) could result in better performance across many dialects simultaneously. These findings are consistent with regression analysis described in the Appendix~\ref{app:grammar_regression}.

Of these rules, we selected three that individually result in high performance degradation, and are obligatory in multiple dialects for further analysis: \emph{existential it}, \emph{drop copula be NP}, and \emph{y'all}. 
On top of each rule, we iteratively apply the other obligatory dialect rules and non-obligatory dialect rules, which---when combined---comprise all dialect rules. We consider only questions where the grammar rule can be applied and are answered correctly in SAE.  Per Figure~\ref{fig:grammar_breakdown}, we find that each of these three rules individually account for at least 45\% of the degradation in performance for these questions compared to SAE. 
This is particularly true for the American dialects (Appalachian English, African American English, and Southern English) where these individual rules account for 64-85\% of overall dialectal degradation.

\section{Discussion}

Our results indicate that LLMs have trouble parsing certain grammatical concepts, leading to fairness concerns for non-``standard'' dialects as quantified by underperformance in knowledge and reasoning benchmark tasks, consistent with findings of underperformance on more explicitly cultural-specific tasks~\cite{ryan2024unintended,hofmann2024ai,lyu2025characterizing}. Our results may be an underestimate of the severity of dialectal underperformance in QA tasks: recent work has found greater performance decreases for human-written AAVE texts relative to Multi-VALUE perturbed texts~\cite{lin2024one}, and multiple choice responses are less variable than open-ended tasks~\cite{hofmann2024ai}.
Our findings could potentially be driven by an underrepresentation of dialectal English in training data;
we hope that pinpointing specific grammatical rules associated with underperformance will allow practitioners to update models for bias mitigation going forward. 

More work needs to be done to explore avenues for such bias mitigation: some researchers have found that in-prompt ``translating''  to SAE still leaves significant performance gaps~\cite{lin2024one}, and others have proposed training LoRAs to map hundreds of grammatical structures to SAE equivalents to mitigate dialectal degradation~\cite{liu2023dada}. 
Going forward, it will be also be important to extend our findings from multiple choice QA tasks to open-ended responses, especially given the increasing concerns of biases in not only valid LLM responses, but also in LLM-generated hallucinations~\cite{huang2025survey,koenecke2024careless}.

Given that the demographics that often use non-``standard'' English have been surveyed to be more likely to rely on LLMs \cite{rainieCloseEncountersAI}, 
it is especially important to ensure that there is a focus on improving LLM performance across dialects---especially from those demographics currently underserved in society and correspondingly underrepresented in training data. 
Looking toward future work in mitigating biases, our findings suggest that practitioners and researchers building LLMs for multi-dialectal users could target future model improvements by training models on QA pair variants---perturbed with even just a handful of important, distinct grammar rules---to yield less disparate performance across multiple dialects.

\section*{Limitations}

We discuss three main limitations of our work.

Firstly, the use of Multi-VALUE as a grammatical translation tool has caveats: the rules are applied following set probabilities of occurrence across different grammatical rules within a dialect, and thus could be debatably similar to a true speaker or writer of that dialect. That said, Multi-VALUE performed human evaluations for a subset of dialects to evaluate their ecological validity~\cite{ziems2022multi}.

Secondly, we make an assumption that grammatical rules would apply, as is, in written text for knowledge and reasoning questions. However, it is possible that speakers of different dialects would typify their grammatical differences in different ways, make different types of errors (such as typos, capitalization, etc.), and so on. While our focus is on dialectal biases, we refer to prior work on the confluence of such biases with typos~\cite{harvey2025framework}---making the case that this intersection likely compounds the degree of underperformance that we found in QA tasks.  

Thirdly, we focus on concerns of LLM underperformance on a subset of QA tasks. Our findings may not generalize to all LLMs (especially newer, larger, or costlier models), nor to all QA tasks---especially for those with open-ended responses, which likely would see lower performance across the board. That said, we hope to raise the point that---while our findings point to LLM underperformance in response to certain grammatical concepts---such underperformance may also be true of gold standards (such as human respondents). For example, negation can be taken at face value, or assumed to be a double negative in some dialectal contexts~\cite{jones2019testifying}---and humans may similarly struggle with such nuanced differentiations if not equipped with the resources to better understand dialectal English. As such, we emphasize the need for human-in-the-loop systems to mitigate both human- and LLM-induced biases by dialect~\cite{alumae2025striving}. 

\section*{Acknowledgments}

This work was supported by a grant from Apple, Inc. Any views, opinions, findings, and conclusions or recommendations expressed in this material are those of the authors and should not be interpreted as reflecting the views, policies or position, either expressed or implied, of Apple Inc.

\bibliography{custom}

\begin{thebibliography}{33}
\providecommand{\natexlab}[1]{#1}

\bibitem[{Alum{\"a}e and Koenecke(2025)}]{alumae2025striving}
Tanel Alum{\"a}e and Allison Koenecke. 2025.
\newblock Striving for open-source and equitable speech-to-speech translation.

\bibitem[{Blodgett et~al.(2016)Blodgett, Green, and O'Connor}]{blodgett_demographic_2016}
Su~Lin Blodgett, Lisa Green, and Brendan O'Connor. 2016.
\newblock \href {https://doi.org/10.18653/v1/D16-1120} {Demographic {Dialectal} {Variation} in {Social} {Media}: {A} {Case} {Study} of {African}-{American} {English}}.
\newblock In \emph{Proceedings of the 2016 {Conference} on {Empirical} {Methods} in {Natural} {Language} {Processing}}, {EMNLP}, pages 1119--1130, Austin, Texas. Association for Computational Linguistics.

\bibitem[{Clark et~al.(2019)Clark, Lee, Chang, Kwiatkowski, Collins, and Toutanova}]{clark2019boolq}
Christopher Clark, Kenton Lee, Ming-Wei Chang, Tom Kwiatkowski, Michael Collins, and Kristina Toutanova. 2019.
\newblock Boolq: Exploring the surprising difficulty of natural yes/no questions.
\newblock \emph{arXiv preprint arXiv:1905.10044}.

\bibitem[{Gao et~al.(2024)Gao, Tow, Abbasi, Biderman, Black, DiPofi, Foster, Golding, Hsu, Le~Noac'h, Li, McDonell, Muennighoff, Ociepa, Phang, Reynolds, Schoelkopf, Skowron, Sutawika, Tang, Thite, Wang, Wang, and Zou}]{eval-harness}
Leo Gao, Jonathan Tow, Baber Abbasi, Stella Biderman, Sid Black, Anthony DiPofi, Charles Foster, Laurence Golding, Jeffrey Hsu, Alain Le~Noac'h, Haonan Li, Kyle McDonell, Niklas Muennighoff, Chris Ociepa, Jason Phang, Laria Reynolds, Hailey Schoelkopf, Aviya Skowron, Lintang Sutawika, and 5 others. 2024.
\newblock \href {https://doi.org/10.5281/zenodo.12608602} {A framework for few-shot language model evaluation}.

\bibitem[{Gil(2003)}]{gil2003english}
David Gil. 2003.
\newblock English goes asian: Number and (in) definiteness in the singlish noun phrase.
\newblock In \emph{Noun phrase structure in the languages of Europe}, pages 467--514. Mouton de Gruyter.

\bibitem[{Harvey et~al.(2025)Harvey, Kizilcec, and Koenecke}]{harvey2025framework}
Emma Harvey, Rene Kizilcec, and Allison Koenecke. 2025.
\newblock \href {https://doi.org/10.1145/3715275.3732137} {A framework for auditing chatbots for dialect-based quality-of-service harms}.
\newblock In \emph{Proceedings of the 2025 ACM Conference on Fairness, Accountability, and Transparency}, FAccT '25, New York, NY, USA. Association for Computing Machinery.

\bibitem[{Hendrycks et~al.(2020)Hendrycks, Burns, Basart, Zou, Mazeika, Song, and Steinhardt}]{hendrycks2020measuring}
Dan Hendrycks, Collin Burns, Steven Basart, Andy Zou, Mantas Mazeika, Dawn Song, and Jacob Steinhardt. 2020.
\newblock Measuring massive multitask language understanding.
\newblock \emph{arXiv preprint arXiv:2009.03300}.

\bibitem[{Hofmann et~al.(2024)Hofmann, Kalluri, Jurafsky, and King}]{hofmann2024ai}
Valentin Hofmann, Pratyusha~Ria Kalluri, Dan Jurafsky, and Sharese King. 2024.
\newblock Ai generates covertly racist decisions about people based on their dialect.
\newblock \emph{Nature}, 633(8028):147--154.

\bibitem[{Huang et~al.(2025)Huang, Yu, Ma, Zhong, Feng, Wang, Chen, Peng, Feng, Qin et~al.}]{huang2025survey}
Lei Huang, Weijiang Yu, Weitao Ma, Weihong Zhong, Zhangyin Feng, Haotian Wang, Qianglong Chen, Weihua Peng, Xiaocheng Feng, Bing Qin, and 1 others. 2025.
\newblock A survey on hallucination in large language models: Principles, taxonomy, challenges, and open questions.
\newblock \emph{ACM Transactions on Information Systems}, 43(2):1--55.

\bibitem[{Jiang et~al.(2023)Jiang, Sablayrolles, Mensch, Bamford, Chaplot, de~las Casas, Bressand, Lengyel, Lample, Saulnier, Lavaud, Lachaux, Stock, Scao, Lavril, Wang, Lacroix, and Sayed}]{jiang2023mistral7b}
Albert~Q. Jiang, Alexandre Sablayrolles, Arthur Mensch, Chris Bamford, Devendra~Singh Chaplot, Diego de~las Casas, Florian Bressand, Gianna Lengyel, Guillaume Lample, Lucile Saulnier, Lélio~Renard Lavaud, Marie-Anne Lachaux, Pierre Stock, Teven~Le Scao, Thibaut Lavril, Thomas Wang, Timothée Lacroix, and William~El Sayed. 2023.
\newblock \href {https://arxiv.org/abs/2310.06825} {Mistral 7b}.
\newblock \emph{Preprint}, arXiv:2310.06825.

\bibitem[{Johnson and VanBrackle(2012)}]{johnson2012linguistic}
David Johnson and Lewis VanBrackle. 2012.
\newblock \href {https://doi.org/10.1016/j.asw.2011.10.001} {Linguistic discrimination in writing assessment: How raters react to african american “errors,” esl errors, and standard english errors on a state-mandated writing exam}.
\newblock \emph{Assessing Writing}, 17(1):35--54.

\bibitem[{Jones et~al.(2019)Jones, Kalbfeld, Hancock, and Clark}]{jones2019testifying}
Taylor Jones, Jessica~Rose Kalbfeld, Ryan Hancock, and Robin Clark. 2019.
\newblock Testifying while black: An experimental study of court reporter accuracy in transcription of african american english.
\newblock \emph{Language}, 95(2):e216--e252.

\bibitem[{Koenecke et~al.(2024)Koenecke, Choi, Mei, Schellmann, and Sloane}]{koenecke2024careless}
Allison Koenecke, Anna Seo~Gyeong Choi, Katelyn~X Mei, Hilke Schellmann, and Mona Sloane. 2024.
\newblock Careless whisper: Speech-to-text hallucination harms.
\newblock In \emph{Proceedings of the 2024 ACM Conference on Fairness, Accountability, and Transparency}, pages 1672--1681.

\bibitem[{Koenecke et~al.(2020)Koenecke, Nam, Lake, Nudell, Quartey, Mengesha, Toups, Rickford, Jurafsky, and Goel}]{koenecke2020racial}
Allison Koenecke, Andrew Nam, Emily Lake, Joe Nudell, Minnie Quartey, Zion Mengesha, Connor Toups, John~R Rickford, Dan Jurafsky, and Sharad Goel. 2020.
\newblock Racial disparities in automated speech recognition.
\newblock \emph{Proceedings of the national academy of sciences}, 117(14):7684--7689.

\bibitem[{Kortmann et~al.(2020)Kortmann, Lunkenheimer, and Ehret}]{ewave}
Bernd Kortmann, Kerstin Lunkenheimer, and Katharina Ehret, editors. 2020.
\newblock \href {https://ewave-atlas.org/} {\emph{eWAVE}}.

\bibitem[{Leimgruber(2013)}]{leimgruber2013trouble}
Jakob~RE Leimgruber. 2013.
\newblock The trouble with world englishes: Rethinking the concept of ‘geographical varieties’ of english.
\newblock \emph{English Today}, 29(3):3--7.

\bibitem[{Lin et~al.(2025)Lin, Mao, La~Malfa, Hofmann, de~Wynter, Wang, Chen, Wooldridge, Pierrehumbert, and Wei}]{lin2024one}
Fangru Lin, Shaoguang Mao, Emanuele La~Malfa, Valentin Hofmann, Adrian de~Wynter, Xun Wang, Si-Qing Chen, Michael~J. Wooldridge, Janet~B. Pierrehumbert, and Furu Wei. 2025.
\newblock \href {https://doi.org/10.18653/v1/2025.acl-long.317} {Assessing dialect fairness and robustness of large language models in reasoning tasks}.
\newblock In \emph{Proceedings of the 63rd Annual Meeting of the Association for Computational Linguistics (Volume 1: Long Papers)}, pages 6317--6342, Vienna, Austria. Association for Computational Linguistics.

\bibitem[{Liu et~al.(2023)Liu, Held, and Yang}]{liu2023dada}
Yanchen Liu, William Held, and Diyi Yang. 2023.
\newblock Dada: Dialect adaptation via dynamic aggregation of linguistic rules.
\newblock \emph{arXiv preprint arXiv:2305.13406}.

\bibitem[{Lyu et~al.(2025)Lyu, Luo, Kang, and Koenecke}]{lyu2025characterizing}
Hanjia Lyu, Jiebo Luo, Jian Kang, and Allison Koenecke. 2025.
\newblock Characterizing bias: Benchmarking large language models in simplified versus traditional chinese.
\newblock In \emph{Proceedings of the 2025 ACM Conference on Fairness, Accountability, and Transparency}, pages 2815--2846.

\bibitem[{Maini et~al.(2024)Maini, Seto, Bai, Grangier, Zhang, and Jaitly}]{maini2024rephrasing}
Pratyush Maini, Skyler Seto, He~Bai, David Grangier, Yizhe Zhang, and Navdeep Jaitly. 2024.
\newblock Rephrasing the web: A recipe for compute and data-efficient language modeling.
\newblock \emph{arXiv preprint arXiv:2401.16380}.

\bibitem[{Martin and Tang(2020)}]{martin2020understanding}
Joshua~L Martin and Kevin Tang. 2020.
\newblock Understanding racial disparities in automatic speech recognition: The case of habitual" be".
\newblock In \emph{Interspeech}, pages 626--630.

\bibitem[{OpenAI(2024)}]{openaiGPT4oMiniAdvancing2024}
OpenAI. 2024.
\newblock \href {https://openai.com/index/gpt-4o-mini-advancing-cost-efficient-intelligence/} {{GPT}-4o mini: advancing cost-efficient intelligence}.

\bibitem[{Penedo et~al.(2024)Penedo, Kydl{\'\i}{\v{c}}ek, allal, Lozhkov, Mitchell, Raffel, Werra, and Wolf}]{penedo2024the}
Guilherme Penedo, Hynek Kydl{\'\i}{\v{c}}ek, Loubna~Ben allal, Anton Lozhkov, Margaret Mitchell, Colin Raffel, Leandro~Von Werra, and Thomas Wolf. 2024.
\newblock \href {https://openreview.net/forum?id=n6SCkn2QaG} {The fineweb datasets: Decanting the web for the finest text data at scale}.
\newblock In \emph{The Thirty-eight Conference on Neural Information Processing Systems Datasets and Benchmarks Track}.

\bibitem[{Rainie(2025)}]{rainieCloseEncountersAI}
Lee Rainie. 2025.
\newblock Close encounters of the {{AI}} kind: {{Main}} report.

\bibitem[{Rickford and Rickford(2007)}]{rickford2007spoken}
John~Russell Rickford and Russell~John Rickford. 2007.
\newblock \emph{Spoken soul: The story of black English}.
\newblock Turner Publishing Company.

\bibitem[{Ryan et~al.(2024)Ryan, Held, and Yang}]{ryan2024unintended}
Michael~J Ryan, William Held, and Diyi Yang. 2024.
\newblock Unintended impacts of llm alignment on global representation.
\newblock \emph{arXiv preprint arXiv:2402.15018}.

\bibitem[{Smitherman(1986)}]{smitherman1986talkin}
G.~Smitherman. 1986.
\newblock \href {https://books.google.com/books?id=HXD7pYv80bUC} {\emph{Talkin and Testifyin: The Language of Black America}}.
\newblock Wayne State University Press.

\bibitem[{Srirag et~al.(2025)Srirag, Sahoo, and Joshi}]{srirag-etal-2025-evaluating}
Dipankar Srirag, Nihar~Ranjan Sahoo, and Aditya Joshi. 2025.
\newblock \href {https://aclanthology.org/2025.sumeval-2.3/} {Evaluating dialect robustness of language models via conversation understanding}.
\newblock In \emph{Proceedings of the Second Workshop on Scaling Up Multilingual {\&} Multi-Cultural Evaluation}, pages 24--38, Abu Dhabi. Association for Computational Linguistics.

\bibitem[{Team et~al.(2024)Team, Mesnard, Hardin, Dadashi, Bhupatiraju, Pathak, Sifre, Rivière, Kale, Love, Tafti, Hussenot, Sessa, Chowdhery, Roberts, Barua, Botev, Castro-Ros, Slone, Héliou, Tacchetti, Bulanova, Paterson, Tsai, Shahriari, Lan, Choquette-Choo, Crepy, Cer, Ippolito, Reid, Buchatskaya, Ni, Noland, Yan, Tucker, Muraru, Rozhdestvenskiy, Michalewski, Tenney, Grishchenko, Austin, Keeling, Labanowski, Lespiau, Stanway, Brennan, Chen, Ferret, Chiu, Mao-Jones, Lee, Yu, Millican, Sjoesund, Lee, Dixon, Reid, Mikuła, Wirth, Sharman, Chinaev, Thain, Bachem, Chang, Wahltinez, Bailey, Michel, Yotov, Chaabouni, Comanescu, Jana, Anil, McIlroy, Liu, Mullins, Smith, Borgeaud, Girgin, Douglas, Pandya, Shakeri, De, Klimenko, Hennigan, Feinberg, Stokowiec, hui Chen, Ahmed, Gong, Warkentin, Peran, Giang, Farabet, Vinyals, Dean, Kavukcuoglu, Hassabis, Ghahramani, Eck, Barral, Pereira, Collins, Joulin, Fiedel, Senter, Andreev, and Kenealy}]{gemmateam2024gemmaopenmodelsbased}
Gemma Team, Thomas Mesnard, Cassidy Hardin, Robert Dadashi, Surya Bhupatiraju, Shreya Pathak, Laurent Sifre, Morgane Rivière, Mihir~Sanjay Kale, Juliette Love, Pouya Tafti, Léonard Hussenot, Pier~Giuseppe Sessa, Aakanksha Chowdhery, Adam Roberts, Aditya Barua, Alex Botev, Alex Castro-Ros, Ambrose Slone, and 89 others. 2024.
\newblock \href {https://arxiv.org/abs/2403.08295} {Gemma: Open models based on gemini research and technology}.
\newblock \emph{arXiv}.

\bibitem[{Welbl et~al.(2017)Welbl, Liu, and Gardner}]{welbl2017crowdsourcing}
Johannes Welbl, Nelson~F Liu, and Matt Gardner. 2017.
\newblock Crowdsourcing multiple choice science questions.
\newblock \emph{arXiv preprint arXiv:1707.06209}.

\bibitem[{Whiteman(2013)}]{whiteman2013dialect}
Marcia~Farr Whiteman. 2013.
\newblock Dialect influence in writing.
\newblock In \emph{Writing}, pages 153--166. Routledge.

\bibitem[{Ziems et~al.(2022{\natexlab{a}})Ziems, Chen, Harris, Anderson, and Yang}]{ziems2022value}
Caleb Ziems, Jiaao Chen, Camille Harris, Jessica Anderson, and Diyi Yang. 2022{\natexlab{a}}.
\newblock Value: Understanding dialect disparity in nlu.
\newblock \emph{arXiv preprint arXiv:2204.03031}.

\bibitem[{Ziems et~al.(2022{\natexlab{b}})Ziems, Held, Yang, Dhamala, Gupta, and Yang}]{ziems2022multi}
Caleb Ziems, William Held, Jingfeng Yang, Jwala Dhamala, Rahul Gupta, and Diyi Yang. 2022{\natexlab{b}}.
\newblock Multi-value: A framework for cross-dialectal english nlp.
\newblock \emph{arXiv preprint arXiv:2212.08011}.

\end{thebibliography}

\appendix

\section{Appendix}
\label{sec:appendix}

\subsection{Grammar Rule Examples}
We provide examples of grammar rule transformations for \emph{existential it}, \emph{y'all}, and \emph{drop copula NP} in Table~\ref{tab:grammar_rule_examples}. We highlight examples that GPT-4o-mini answered correctly in SAE but incorrectly after grammatical perturbation. Due to generally high performance in SciQ, no such examples existed for existential it.

\subsection{Grammar Rule Regression Analysis}\label{app:grammar_regression}
Overall, the results presented in Figures~\ref{fig:grammar_bars} and \ref{fig:grammar_breakdown} are consistent with regression analyses spanning LLMs. Specifically, we run a logistic regression where the outcome is a binary for whether an individual grammatically-perturbed question was answered correctly or incorrectly, and covariates include binary variables for whether a grammar rule category is perturbed in that question text, a binary indicator for whether the original SAE question was answered correctly, and binary indicators for the LLM. We run this regression on 535,239 samples (i.e., the sum of all questions in three datasets with at least one rule applied, times 6 dialects times 3 LLMs), and find that 12 out of 13 grammatical rule categories have a negative effect on accuracy and are statistically significant at the 0.05 level. Results are displayed in Table~\ref{tab:logit_regression}. 
The large negative coefficient for the grammatical categorty of \emph{Agreement} is consistent with our findings for \emph{existential it} and \emph{drop copula be NP}, which are contained in that category. 

\subsection{Other Data Subsetting Variants}\label{app:other_metric_variants}
Table~\ref{tab:dialect_counts_only_applied_sae_correct} reflects the counts of the number of questions answered correctly under the settings indicated by Table~\ref{tab:dialect_accuracy_only_applied}. To account for the inclusion of unperturbed questions, we generate results analogous to Tables~\ref{tab:dialect_accuracy_only_applied} and \ref{tab:dialect_accuracy_only_applied_sae_correct} when including unperturbed questions in each of the dialects; these results (showing similar degradation across dialects) are reflected in Tables~\ref{tab:dialect_accuracy_sae_correct} and ~\ref{tab:dialect_accuracy}, respectively.

\subsection{Singaporean English Case Study}\label{app:singe}

Singaporean English (CollSgE) exhibits the most substantial performance degradation among the dialects examined. This pronounced underperfomance likely stems from its distinctive status as a contact variety with strong pidgin/creole characteristics that fundamentally differentiate its structure from SAE \cite{leimgruber2013trouble, gil2003english}. Unlike other non-standard varieties that share more grammatical patterns with SAE, Singlish employs multiple syntactic structures that systematically diverge from SAE. The concentration of these high-impact grammatical features within Singaporean English may explain why models trained predominantly on Western varieties struggle disproportionately with this dialect compared to others.

We explore how interactions between individual grammar rules affect performance degradation. 
We use Singaporean English as a case study given its consistently high performance degradation. We focus on \emph{null prepositions}, \emph{one relativizer} and \emph{drop copula be NP} as rules that both individually have a significant impact and regularly co-occur. To do this, we examine the effect of iteratively applying each of the three grammatical rules to the original dataset. We consider the subset of questions where all three of those rules can be applied. We compare the expected performance degradation of simply adding individual rule degradations to observed degradations in Tables~\ref{tab:singe_interaction} and \ref{tab:singe_interaction_sae_corr}. 
We find that while there is additional performance reduction when applying multiple rules on questions originally answerable in SAE, the degradation is different than adding their individual performance reductions together, pointing towards an interaction effect among co-occurring grammar rules.

\subsection{Licensing and Code}

We adhere to  the licensing requirements and intended usage of the datasets, models, and packages used. We release our code under the MIT License at \url{https://github.com/peridotleaves/Dialect_Bias}.

\begin{table*}[ht]
\centering
\small
\begin{tabular}{l|p{3cm}|p{2cm}|l|p{3cm}|p{2cm}}
\hline
Dataset & Original & Answer & Rule & Transformed & Answer to Transformed \\
\hline
boolq & is there a difference between maid of honour and chief bridesmaid & no & existential\_it & is it a difference between maid of honour and chief bridesmaid & yes \\
sciq & Which disease occurs when there is not enough hemoglobin in the blood? & anemia & existential\_it & Which disease occurs when it is not enough hemoglobin in the blood? & anemia \\
mmlu & How many kcal are there in one gram of ethanol? & 29.7 kJ or 7.1 kcal per g & existential\_it & How many kcal is it in one gram of ethanol? & 36.5 kJ or 8.1 kcal per g \\
boolq & can you drive with a beer in texas & no & yall & can y'all drive with a beer in texas & yes \\
sciq & What should you use to protect your eyes from chemicals? & eye goggles & yall & What should y'all use to protect your eyes from chemicals? & certain goggles \\
mmlu & You need to construct a 94\% confidence interval for a population proportion. What is the upper critical value of z to be used in constructing this interval? & 1.88 & yall & Y'all gotta construct a 94\% confidence interval for a population proportion. What is the upper critical value of z to be used in constructing this interval? & 1.96 \\
boolq & pecans and walnuts in the same family & yes & drop\_copula\_np & are pecans and walnuts in the same family & no \\
sciq & Alpha emission is a type of what? & radioactivity & drop\_copula\_np & Alpha emission a type of what? & radiation \\
mmlu & Which fraction is greater than 2 over 5? & 5 over 10 & drop\_copula\_np & Which fraction greater than 2 over 5? & 4 over 10 \\
\hline
\end{tabular}
\caption{Examples of Grammar Transformations with Corresponding Answers}
\label{tab:grammar_rule_examples}
\end{table*}

\begin{figure*}[ht]
  \centering
  \begin{subfigure}[t]{0.49\linewidth}
    \centering
    \includegraphics[width=\linewidth]{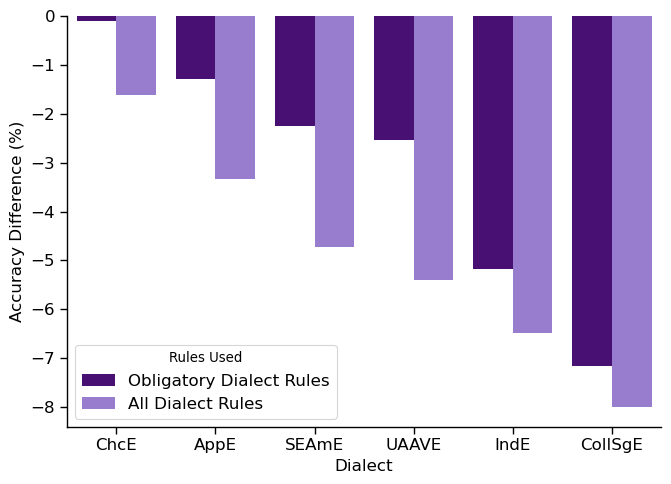}
    \subcaption{Full dataset (including unperturbed questions).}
    \label{fig:oblig-b}
  \end{subfigure}
  \hfill
    \begin{subfigure}[t]{0.49\linewidth}
    \centering
    \includegraphics[width=\linewidth]{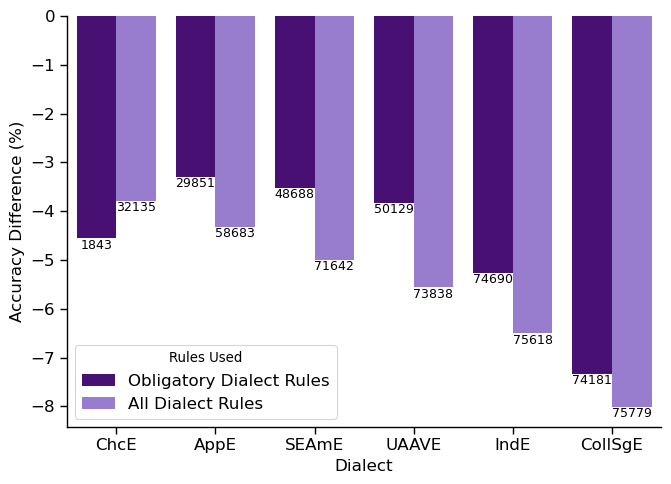}
    \subcaption{Restricted to question subset where at least one grammar is rule applied. Numbers along bars refer to number of questions across datasets.}
    \label{fig:oblig-a}
  \end{subfigure}
  \caption{Breakdown of the extent to which overall accuracy decreases can be attributed to obligatory grammatical rules compared to all dialect rules. Abbreviations are: African American English (UAAVE), Singaporean English (CollSgE), Indian English (IndE), Appalachian English (AppE), Chicano English (ChcE), and Southern English (SEAmE).}
  \label{fig:oblig_vs_all}
\end{figure*}

\begin{table*}[ht]
\scriptsize
\setlength{\tabcolsep}{4pt}
\centering
\begin{tabular}{l|ccc|ccc|ccc|}
\hline
\multirow{2}{*}{English Variety} & \multicolumn{3}{c|}{BoolQ Accuracy} & \multicolumn{3}{c|}{SciQ Accuracy} & \multicolumn{3}{c|}{MMLU Accuracy}  \\
\cline{2-10}
 & Gemma 2B & Mistral 7B & GPT4o-mini & Gemma 2B & Mistral 7B & GPT4o-mini & Gemma 2B & Mistral 7B & GPT4o-mini \\ 
\hline
Standard American English  & 6670 & 7982  & 8202  & 10983 & 11256 & 11379 & 3974 & 7150 & 8344 \\
Chicano English      & 2667 & 3142 & 3228 & 4077 & 4186 & 4229 & 2221 & 3879 & 4506 \\
Appalachian English    & 4156 & 4906 & 5028 & 9317 & 9530 & 9631 & 3352 & 5930 & 6833 \\
Southern English     & 5925 & 7090 & 7278 & 10706 & 10973 & 11094 & 3797 & 6817 & 7962 \\
African American English     & 6477 & 7755 & 7962 & 10798 & 11065 & 11183 & 3808 & 6837 & 7953 \\
Indian English      & 6650 & 7957 & 8175 & 10970 & 11241 & 11364 & 3933 & 7072 & 8256 \\
Singaporean English   & 6655 & 7960 & 8181 & 10979 & 11252 & 11374 & 3957 & 7118 & 8303 \\
\hline
\end{tabular}
\caption{Counts of Standard American English and English varieties' questions (grammatically unperturbed questions excluded) answered correctly.}
\label{tab:dialect_counts_only_applied_sae_correct}
\end{table*}

\begin{table*}[ht]
\scriptsize
\setlength{\tabcolsep}{4pt} %
\centering
\begin{tabular}{l|ccc|ccc|ccc|}
\hline
\multirow{2}{*}{English Variety} & \multicolumn{3}{c|}{BoolQ Accuracy (\%)} & \multicolumn{3}{c|}{SciQ Accuracy (\%)} & \multicolumn{3}{c|}{MMLU Accuracy (\%)}  \\
\cline{2-10}
 & Gemma 2B & Mistral 7B & GPT4o-mini & Gemma 2B & Mistral 7B & GPT4o-mini & Gemma 2B & Mistral 7B & GPT4o-mini \\
\hline
Standard American English & 71.3 & 85.4 & 87.7 & 94.3 & 96.6 & 97.7 & 34.1 & 61.3 & 71.5 \\
Chicano English & 70.9 (-0.4) & 84.8 (-0.6) & 87.3 (-0.4) & 94.1 (-0.2) & 96.6 (0.0) & 97.6 (-0.1) & 34.1 (0.0) & 61.1 (-0.3) & 70.8 (-0.7) \\
Appalachian English & 70.3 (-1.0) & 83.9 (-1.5) & 86.4 (-1.3) & 93.3 (-1.0) & 96.4 (-0.2) & 97.2 (-0.5) & 34.5 (+0.4) & 60.3 (-1.0) & 70.5 (-1.0) \\
Southern English & 69.3 (-2.0) & 83.3 (-2.1) & 85.9 (-1.8) & 93.5 (-0.8) & 96.3 (-0.3) & 97.2 (-0.5) & 34.1 (0.0) & 59.9 (-1.4) & 70.2 (-1.3) \\
African American English & 67.8 (-3.5) & 82.7 (-2.8) & 85.9 (-1.8) & 93.3 (-1.0) & 96.1 (-0.5) & 97.0 (-0.7) & 34.2 (+0.1) & 59.7 (-1.6) & 69.7 (-1.8) \\
Indian English & 68.1 (-3.2) & 81.2 (-4.2) & 85.4 (-2.3) & 92.7 (-1.6) & 95.8 (-0.8) & 96.5 (-1.2) & 34.7 (+0.6) & 59.4 (-1.9) & 68.9 (-2.6) \\
Singaporean English & 66.5 (-4.8) & 79.8 (-5.6) & 84.6 (-3.1) & 91.7 (-2.6) & 94.7 (-1.9) & 96.1 (-1.6) & 34.1 (0.0) & 58.5 (-2.8) & 68.7 (-2.8) \\
\hline
\end{tabular}
\caption{Performance comparison across English varieties (averaged over all questions), with unperturbed questions included (in contrast to Table~\ref{tab:dialect_accuracy_only_applied}).}
\label{tab:dialect_accuracy}
\end{table*}

\begin{table*}[ht]
\scriptsize
\setlength{\tabcolsep}{3pt} %
\centering
\begin{tabular}{l|ccc|ccc|ccc|}
\hline
\multirow{2}{*}{English Variety} & \multicolumn{3}{c|}{BoolQ Accuracy (\%)} & \multicolumn{3}{c|}{SciQ Accuracy (\%)} & \multicolumn{3}{c|}{MMLU Accuracy (\%)}  \\
\cline{2-10}
 & Gemma 2B  & Mistral 7B & GPT4o-mini & Gemma 2B & Mistral 7B & GPT4o-mini & Gemma 2B & Mistral 7B & GPT4o-mini \\
 
  & 6670/9348  & 7982/9348  & 8202/9348  & 10983/11647 & 11256/11647 & 11379/11647 & 3974/11672 & 7150/11672 & 8344/11672   \\
\hline
Standard American English 
& 100 & 100 & 100 
& 100 & 100 & 100 
& 100 & 100 & 100 \\
Chicano English   
& 97.5 (-2.5) & 98.3 (-1.7) & 98.7 (-1.3) 
& 99.7 (-0.3) & 99.9 (-0.1) & 99.8 (-0.2) 
& 94.0 (-6.0) & 96.2 (-3.8) & 97.4 (-2.6) \\
Appalachian English    
& 95.0 (-5.0) & 96.1 (-3.9) & 96.8 (-3.2) 
& 98.4 (-1.6) & 99.2 (-0.8) & 99.3 (-0.7) 
& 88.9 (-11.1) & 94.2 (-5.8) & 94.9 (-5.1) \\
Southern English    
& 91.2 (-8.8) & 93.9 (-6.1) & 95.4 (-4.6) 
& 98.5 (-1.5) & 99.1 (-0.9) & 99.0 (-1.0) 
& 83.8 (-16.2) & 92.9 (-7.1) & 92.7 (-7.3) \\
African American English    
& 86.3 (-13.7) & 92.1 (-7.9) & 95.2 (-4.8) 
& 98.2 (-1.8) & 99.1 (-0.9) & 98.9 (-1.1) 
& 85.1 (-14.9) & 92.7 (-7.3) & 92.6 (-7.4) \\
Indian English     
& 86.9 (-13.1) & 90.2 (-9.8) & 93.6 (-6.4) 
& 97.5 (-2.5) & 98.4 (-1.6) & 98.5 (-1.5) 
& 81.5 (-18.5) & 91.3 (-8.7) & 90.9 (-9.1) \\
Singaporean English  
& 83.4 (-16.6) & 88.2 (-11.8) & 92.4 (-7.6) 
& 96.4 (-3.6) & 98.0 (-2.0) & 97.4 (-2.6) 
& 78.5 (-21.5) & 90.0 (-10.0) & 88.8 (-11.2) \\
\hline
\end{tabular}
\caption{Performance comparison across English varieties, with unperturbed questions included (in contrast to Table~\ref{tab:dialect_accuracy_only_applied_sae_correct}), conditioned on SAE responses being correct.}
\label{tab:dialect_accuracy_sae_correct}
\end{table*}

\begin{table*}[htbp]
\centering
\begin{tabular}{lrrrr}
\hline
\textbf{Feature} & \textbf{Coefficient} & \textbf{Std. Error} & \textbf{z-value} & \textbf{p-value} \\
\hline
Constant & 0.8441 & 0.009 & 95.915 & 0.000*** \\
Pronouns & -0.1483 & 0.010 & -15.345 & 0.000*** \\
Noun Phrase & -0.1037 & 0.008 & -12.900 & 0.000*** \\
Tense+Aspect & -0.1571 & 0.009 & -17.650 & 0.000*** \\
Modal Verbs & -0.0802 & 0.012 & -6.769 & 0.000*** \\
Verb Morphology & -0.1029 & 0.010 & -10.411 & 0.000*** \\
Negation & -0.0996 & 0.015 & -6.645 & 0.000*** \\
Agreement & -0.1216 & 0.009 & -14.035 & 0.000*** \\
Relativization & -0.1434 & 0.010 & -14.126 & 0.000*** \\
Complementation & -0.0208 & 0.013 & -1.567 & 0.117 \\
Adverb Subordination & -0.1932 & 0.026 & -7.368 & 0.000*** \\
Adverbs+Prepositions & -0.0361 & 0.010 & -3.610 & 0.000*** \\
Discourse+Word Order & -0.0992 & 0.008 & -11.809 & 0.000*** \\
LLM (GPT4o-mini) & 1.2021 & 0.009 & 132.519 & 0.000*** \\
LLM (Mistral 7B) & 0.8494 & 0.009 & 98.638 & 0.000*** \\
Dataset (MMLU) & -1.1589 & 0.009 & -126.680 & 0.000*** \\
Dataset (SciQ) & 1.8128 & 0.013 & 136.216 & 0.000*** \\
SAE Accuracy & 0.2995 & 0.009 & 33.896 & 0.000*** \\
\hline
\multicolumn{5}{l}{\small Observations: 535239 \quad Pseudo $R^2$: 0.2012 \quad Log-Likelihood: $-2.3654 \times 10^5$} \\
\multicolumn{5}{l}{\small $^a$LLM (Gemma) used as reference category \quad $^b$Dataset (boolq) used as reference category} \\
\multicolumn{5}{l}{\small ***p<0.001, **p<0.01, *p<0.05} \\
\end{tabular}
\caption{Logistic Regression of Linguistic Features on Question-Level Accuracy.}
\label{tab:logit_regression}
\end{table*}

\begin{table*}[htbp]
\centering
\resizebox{\textwidth}{!}{%
\begin{tabular}{p{4.5cm}p{1.75cm}lrp{1.5cm}p{4.25cm}}
\hline
\textbf{Grammar Rule} & \textbf{Accuracy Diff. (\%)} & \textbf{p-value}  & \textbf{Count} & \textbf{Dialect} & \textbf{eWAVE Rule \# and Name} \\
\hline
existential\_it & -3.11 & 9.92e-12 & 1629 & UAAVE, AppE, CollSgE & 173. Variant forms of dummy subject \textit{there} in existential clauses \\
invariant\_tag\_can\_or\_not & -3.01 & 4.23e-18 & 3759 & CollSgE & 166. Invariant tag \textit{can or not?} \\
give\_passive & -2.68 & 1.79e-05 & 895 & CollSgE & 153. \textit{Give} passive: NP1 (patient) + \textit{give} + NP2 (agent) + V \\
invariant\_tag\_non\_concord & -2.25 & 1.51e-11 & 3759 & IndE, CollSgE & 165. Invariant non-concord tags \\
completive\_finish & -1.94 & 2.43e-04 & 739 & CollSgE &  110.\textit{Finish}-derived completive markers \\
fronting\_pobj & -1.77 & 2.50e-46 & 14760 & IndE & 224. Other possibilities for fronting than SAE \\
yall & -1.26 & 0.0016 & 1377 & UAAVE, AppE, SEAmE & 34. Forms or phrases for the second person plural pronoun other than \textit{you} \\
null\_prepositions & -1.22 & 1.08e-47 & 28036 & CollSgE & Omission of SAE prepositions \\
aint\_be & -1.13 & 0.0288 & 882 & UAAVE, AppE, SEAmE & 155. \textit{Ain’t} as the negated form of \textit{be} \\
existential\_got & -1.09 & 0.0086 & 1525 & CollSgE & 205. Existentials with forms of \textit{get} \\
drop\_copula\_be\_NP & -0.96 & 9.52e-07 & 5911 & UAAVE, CollSgE & 176. Deletion of copula \textit{be} before NPs \\
definite\_for\_indefinite\_articles & -0.61 & 2.80e-08 & 11881 & IndE & 60. Use of definite article where SAE has indefinite article \\
one\_relativizer & -0.50 & 1.83e-09 & 21879 & CollSgE & 195. Postposed \textit{one} as sole relativizer \\
drop\_aux\_yn & -0.48 & 2.90e-06 & 12960 & IndE, CollSgE & 229. No inversion/no auxiliaries in main clause yes/no questions \\
remove\_det\_indefinite & -0.33 & 0.0059 & 9671 & IndE, CollSgE & 63. Zero article used where SAE has indefinite article \\
progressives & -0.33 & 0.0018 & 12400 & SEAmE, IndE & 88. Wider range of uses of progressive \textit{be + V-ing} than in SAE \\
mass\_noun\_plurals & -0.29 & 0.0015 & 15760 & IndE & 55. Different count/mass noun distinctions: plural for SAE singular \\
zero\_plural & -0.27 & 0.0011 & 16804 & CollSgE & 58. Optional plural marking for non-human nouns \\
drop\_aux\_wh & -0.27 & 0.0047 & 10926 & IndE, CollSgE & 228. No inversion/no auxiliaries in wh-questions \\
\hline
\end{tabular}
}
\caption{Obligatory Grammar Rules with Statistically Significant Accuracy Decreases from SAE. We report accuracy differences for the subset of questions the grammar rule can be applied. Abbreviations are: African American English (UAAVE), Singaporean English (CollSgE), Indian English (IndE), Appalachian English (AppE), and Southern English (SEAmE). Rule definitions are eWAVE feature names \cite{ewave}.}
\label{tab:grammar_table}
\end{table*}

\begin{table*}[htbp]
\centering
\setlength{\tabcolsep}{4pt}
\resizebox{\textwidth}{!}{%
\begin{tabular}{l|c|c|c}
\hline
\textbf{Grammar Rules} & \textbf{Accuracy Diff. (\%)} & \textbf{Additive Diff. (\%)} & \textbf{Interaction Gap (\%)} \\
\hline
null\_prepositions & -0.49 & — & — \\
drop\_copula\_be\_NP & -0.34 & — & — \\
one\_relativizer & -0.38 & — & — \\
\hline
null\_prepositions + drop\_copula\_be\_NP & -1.44 & -0.83 & -0.61 \\
null\_prepositions + one\_relativizer & -1.10 & -0.88 & -0.22 \\
All three rules combined & -2.00 & -1.21 & -0.79 \\
\hline
\end{tabular}
}
\caption{Cumulative interaction effect for Singaporean English for three commonly co-occurring rules (null prepositions, drop copula be NP, and one relativizer). The set of questions used to generate averages is restricted to the for which all three grammar rules could be applied (n=4452). The Additive Diff column refers to naively summing the accuracy differences from individual accuracies (e.g., the additive difference for null prepositions occurring and drop copula occurring is -0.49-0.34=-.83). The Interaction Gap column refers to the difference between these naive sums, versus those found in the Accuracy Diff column (which instead reflect questions in which null prepositions and drop copula co-occur), e.g. -1.44-(-0.83)=-0.61. This reveals an interaction effect in grammar rules co-occurring.}
\label{tab:singe_interaction}
\end{table*}

\begin{table*}[htbp]
\centering
\setlength{\tabcolsep}{4pt}
\resizebox{\textwidth}{!}{%
\begin{tabular}{l|c|c|c}
\hline
\textbf{Grammar Rules} & \textbf{Accuracy Diff. (\%)} & \textbf{Additive Diff. (\%)} & \textbf{Interaction Gap (\%)} \\
\hline
null\_prepositions & -2.79 & — & — \\
drop\_copula\_be\_NP & -2.95 & — & — \\
one\_relativizer & -2.51 & — & — \\
\hline
null\_prepositions + drop\_copula\_be\_NP & -4.72 & -5.74 & +1.02 \\
null\_prepositions + one\_relativizer & -4.25 & -5.30 & +1.05 \\
All three rules combined & -5.77 & -8.25 & +2.48 \\
\hline
\end{tabular}
}
\caption{Cumulative interaction effect for Singaporean English for three commonly co-occurring rules (null prepositions, drop copula be NP, and one relativizer). The set of questions used to generate averages is restricted to those for which the SAE question variant was answered correctly, and all three grammar rules could be applied (n=3623). The Additive Diff column refers to naively summing the accuracy differences from individual accuracies (e.g., the additive difference for null prepositions occurring and drop copula occurring is -2.79-2.95=-5.74). The Interaction Gap column refers to the difference between these naive sums, versus those found in the Accuracy Diff column (which instead reflect questions in which null prepositions and drop copula co-occur), e.g. -4.72-(-5.74)=+1.02. This reveals an interaction effect in grammar rules co-occurring.}
\label{tab:singe_interaction_sae_corr}
\end{table*}

\begin{table*}[ht]
\small
\centering
\resizebox{\textwidth}{!}{%
\begin{tabular}{lllccc}
\hline
Benchmark & English Variety & Original PPL & Dialect PPL & PPL Difference & PPL Increase Percentage (\%) \\
\hline
SCIQ & African American English & 46.52 & 240.57 & 194.05 & 558.97 \\
 & Appalachian English & 46.45 & 153.28 & 106.83 & 303.52 \\
 & Chicano English & 47.14 & 94.36 & 47.22 & 103.99 \\
 & Indian English & 46.91 & 492.89 & 445.98 & 1362.89 \\
 & Singaporean English & 46.93 & 1196.28 & 1149.35 & 3517.43 \\
 & Southern English & 46.52 & 223.88 & 177.36 & 510.67 \\
\hline
BOOLQ & African American English & 399.14 & 803.36 & 404.22 & 247.25 \\
 & Appalachian English & 399.14 & 622.58 & 223.44 & 144.59 \\
 & Chicano English & 355.87 & 545.51 & 189.64 & 121.02 \\
 & Indian English & 399.14 & 2167.69 & 1768.54 & 948.57 \\
 & Singaporean English & 399.14 & 5109.68 & 4710.54 & 2713.13 \\
 & Southern English & 399.14 & 865.94 & 466.80 & 288.51 \\
\hline
MMLU & African American English & 77.19 & 260.47 & 183.28 & 544.61 \\
 & Appalachian English & 36.75 & 127.98 & 91.23 & 273.49 \\
 & Chicano English & 30.65 & 77.05 & 46.41 & 138.91 \\
 & Indian English & 78.97 & 438.70 & 359.73 & 1434.15 \\
 & Singaporean English & 78.72 & 959.37 & 880.65 & 3318.71 \\
 & Southern English & 75.95 & 196.08 & 120.13 & 485.55 \\
\hline
\end{tabular}
}
\caption{Perplexity Analysis by Dialect and Benchmark}
\label{tab:perplexity_dialects}
\end{table*}

\end{document}